\pdfoutput=1

\documentclass[11pt]{article}

\usepackage{acl}

\usepackage{times}
\usepackage{latexsym}

\usepackage[T1]{fontenc}

\usepackage[utf8]{inputenc}

\usepackage{microtype}

\usepackage{inconsolata}

\usepackage{multirow}
\usepackage{graphicx}
\usepackage{subcaption}
\usepackage{amsmath}
\usepackage{booktabs}
\usepackage{listings}
\usepackage{xcolor}
\usepackage{tabularx}
\usepackage{booktabs}
\usepackage{colortbl}

\lstdefinestyle{customPromptStyle}{
    basicstyle=\ttfamily\small,
    moredelim=**[is][\color{blue}\bfseries]{<}{>},
    moredelim=**[is][\color{black}]{@}{@}, 
    breaklines=true, 
    captionpos=b, 
    frame=tb, 
}

\title{Speaker Verification in Agent-generated Conversations}

\author{Yizhe Yang${}^{1,2}$~\thanks{Work was done during a visit at SMU.},~~Palakorn Achananuparp${}^{3}$,~~Heyan Huang${}^{1,2}$~\thanks{Corresponding Author},~~Jing Jiang${}^{3}$,~~Ee-Peng Lim${}^{3}$ \\
        ${}^{1}$School of Computer Science and Technology, Beijing Institute of Technology\\
        ${}^{2}$Southeast Academy of Information Technology, Beijing Institute of Technology\\
        ${}^{3}$Singapore Management University\\
        \texttt{\{yizheyang,hhy63\}@bit.edu.cn, \{palakorna,jingjiang,eplim\}@smu.edu.sg}}

\begin{document}
\maketitle
\begin{abstract}

The recent success of large language models (LLMs) has attracted widespread interest to develop role-playing conversational agents personalized to the characteristics and styles of different speakers to enhance their abilities to perform both general and special purpose dialogue tasks. However, the ability to personalize the generated utterances to speakers, whether conducted by human or LLM, has not been well studied. To bridge this gap, our study introduces a novel evaluation challenge: speaker verification in agent-generated conversations, which aimed to verify whether two sets of utterances originate from the same speaker. To this end, we assemble a large dataset collection encompassing thousands of speakers and their utterances. We also develop and evaluate speaker verification models under experiment setups. We further utilize the speaker verification models to evaluate the personalization abilities of LLM-based role-playing models. Comprehensive experiments suggest that the current role-playing models fail in accurately mimicking speakers, primarily due to their inherent linguistic characteristics.

\end{abstract}

\section{Introduction}

The recent advances in large language models (LLMs) have significantly increased the capabilities of conversational AI to solve challenging dialogue problems~\citep{zhao2023survey,chang2023survey,park2023generative,wang2023survey,gao2023s}. In particular, LLM-based role-playing chatbots have been developed to simulate speakers of different personal attributes and linguistic styles so as to provide more immersive interaction with users~\citep{shanahan2023role}. 
Nevertheless, there is a lack of study on how well the agent-generated utterances are personalized according to the target characters/speakers.  Conventional evaluations of role-playing agent models typically focus on assessing their background knowledge through question answering, or rely on judgement by humans or other LLMs~\citep{wang2023rolellm,tu2024charactereval,shenRoleEvalBilingualRole2023,xiaoHowFarAre2023}. These evaluation methodologies do not address the consistency between utterances of the simulated speaker and that of the target speaker in linguistic style and personal characteristics. To bridge this gap, we introduce the speaker verification task in agent-based conversations as an important approach to evaluate the personalization ability of conversational agents.

Speaker verification refers to the task of determining if two sets of utterances belong to the same speaker. In the context of evaluating role-playing agent models, a positive match between the generated and real utterances of a speaker suggests the former preserves the speaker's distinct linguistic style and personal traits. Unlike the authorship attribution task~\citep{rivera-sotoLearningUniversalAuthorship2021,wegmannSameAuthorJust2022}, speaker verification may involve two sets of utterances from unseen speakers (i.e., speakers not seen in the training data). Given that role-playing agents can simulate virtual characters or user-customized personas in diverse conversation settings (e.g., movies, sitcoms, interest sharing, and counselling), we aim to develop models that can robustly verify utterances of unseen speakers in these different settings. 
While authorship attribution is closely related, speaker verification goes beyond linguistic style to also consider content, which can reveal personal characteristics, including personality traits, moral foundations, value.
To evaluate the speaker verification models, we compile conversation data from various sources, encompassing thousands of speakers and employ a variety of methods including style embedding, authorship verification, and fine-tuned models. 

Following previous research that highlights the significant influence of topic on style identification~\citep{wegmannSameAuthorJust2022}, we have designed our evaluation to control for conversation setting by introducing three levels of difficulty: {\bf Base}, where the two sets of utterances may come from different conversation contexts and thus easier to distinguish; {\bf Hard}, where utterances are from the same conversation context; and {\bf Harder}, where utterances are from the same conversation.  This approach allows us to isolate the impact of conversation topic on speaker verification accuracy.
Moreover, Our experiment design includes diverse testing scenarios of different degrees of exposure to speakers and utterances, corresponding to different applications. Our extensive experiment results show that neither non-experts nor the ChatGPT are able to perform speaker verification accurately, highlighting both the challenges and the limitations of current evaluation. In contrast, our proposed fine-tuned models demonstrate the ability to differentiate between speakers effectively, and thus the potential to evaluate the ability of role-playing agents personalizing utterances to speakers. Despite the poorer performance at the Harder level, we argue that the topic should also be considered in speaker verification processes, as it reflects the speaker's personal topic preference. For example, when agents simulate Harry Potter, the topic should be around the wizarding world, reflecting his magical background. In contrast, the topic for simulating Sheldon Cooper, a physicist, may focus on theoretical physics, highlighting their distinct backgrounds and interests. 



Moreover, we utilize the developed speaker verification models to evaluate how well agent-generated utterances could preserve the personal identity of the speakers. We expect that the utterances generated by the role-playing model should closely mirror the style and persona of the speaker being simulated, while also demonstrating clear differentiation from other speakers, including the model itself. To evaluate these two aspects, we introduce two metrics: (a) {\em Simulation Score}, which evaluates the alignment of agent-generated utterances with actual utterances from the target speaker, and (b) {\em Distinction Score}, which measures the differences among utterances generated by the same agent model when simulating various speakers. Moreover, we visualize the similarity distributions estimated by speaker verification models. The separability between similarity distributions of positive and negative pairs serves as an indicator of the model's simulation proficiency. Our analyses reveal that LLMs, whether prompted or specifically fine-tuned, generally struggle to simulate the style or personal characteristics of a target speaker. Furthermore, the limited variability in utterances generated by the same role-playing model for different speakers points to an inherent linguistic consistency within these LLMs. This consistency limits the models' ability to significantly alter their linguistic style when tasked for diverse speakers.

We summarize our contributions as follows:
\begin{itemize}
    \item We define a novel task of speaker verification in agent-based conversations, which is essential to evaluate the role-playing models. To develop our models, we compiled a dataset from a wide range of conversations involving thousands of speakers. Our study indicates that specifically fine-tuned models demonstrate the most promising performance in speaker verification tasks. 
    \item We utilize the developed speaker verification models to assess current role-playing models, addressing a gap in the evaluation. Our analyses reveals that current LLM-based role-playing models fail to simulate the target speaker and exhibit a specific linguistic style that proves difficult to modify. Our research introduces a rigorous evaluation metric for role-playing models, highlighting a substantial opportunity for improvement in this area. 
\end{itemize}

The structure of our paper is organized as follows: After introducing related work (Section~\ref{sec:related work}), we first construct a dataset for the task of speaker verification in conversations (Section~\ref{sec: data collection}). We then test the performance on this task by humans and by ChatGPT (Section~\ref{sec:human}). Observing that neither humans nor LLMs can perform well, we design several speaker verification models and evaluate their performance on the dataset we have constructed (Section~\ref{sec:model}). Finally, we use our trained speaker verification model to evaluate several existing role-playing models (Section~\ref{sec:evalaution}). Our code and data are available for public access to facilitate further research.\footnote{The source code and datasets associated with this study can be accessed at \url{https://github.com/IzzetYoung/AGCSpeakerVerification}.}

\section{Related Work}
\label{sec:related work}

\subsection{Speaker Verification in Speech Processing}

Speaker verification in speech processing seeks to confirm if a voice claim matches the true identity of the speaker by comparing the voice sample against a pre-stored voiceprint. In this topic, researchers have developed methods based on neural network structures and methodologies, often leveraging both publicly available and proprietary datasets~\citep{wan2018generalized,li2021fdn,liu2023cross,kim2023analysis}. Unlike the above works, our study focuses on speaker verification using conversation data and considers more evaluation settings which have not been studied, e.g., verification involving completely unseen speakers.
For example, FDN~\citep{li2021fdn} details an innovative network that layers intonation analysis atop lower-level voice embeddings, markedly boosting system precision, a claim substantiated by tests on the VoxCeleb1 dataset. Moreover, research by \citet{liu2023cross} investigates the integration of visual cues like lip movements with auditory speech, establishing a novel dual-mode learning approach in speech processing. 

\subsection{Author Verification}

Authorship verification aims to identify whether two pieces of text are written by the same person. For this task, researchers have proposed contrastive learning to cluster the representations of texts from the same author closer to one another while increasing the distance between text representations of different authors~\citep{rivera-sotoLearningUniversalAuthorship2021,reimers2019sentence,wangCanAuthorshipRepresentation2023,wegmannSameAuthorJust2022}. 
Recently, \citet{aggazzottiCanAuthorshipAttribution2023} evaluate authorship attribution models's capacity to identify speakers in speech transcripts, a task similar to ours. Our study focus on verifying utterances from unseen speakers in various conversations like movies, sitcoms, interest sharing, and counselling, beyond mere speech transcripts. Additionally, we utilize speaker verification models to assess the role-playing models, and introduce a novel framework for evaluation.


\subsection{Role-Playing Conversation AI}

Recent research in LLMs has focused on exploring their potential as role-playing agents through strategies like prompting or fine-tuning~\citep{shaoCharacterLLMTrainableAgent2023,wang2023rolellm,zhouCharacterGLMCustomizingChinese2023}. Nevertheless, the evaluation of generated utterances with respect to the target speaker remains under-explored. Prior research typically evaluates role-playing agents in two main approaches: (1) through evaluation judgements made by humans or LLMs~\citep{shaoCharacterLLMTrainableAgent2023,wang2023rolellm,zhouCharacterGLMCustomizingChinese2023}, and (2) by employing question-answering tests or reward models on specific benchmarks~\citep{tu2024charactereval,shenRoleEvalBilingualRole2023,xiaoHowFarAre2023}. Nevertheless, our research highlights a significant gap in the ability of non-experts and LLMs to distinguish between different speakers, indicating their unreliability in the evaluation. Moreover, existing benchmarks concentrate on a limited set of speakers and the question-answering testing is mismatch with conversation. Consequently, we propose speaker verification models to assess the extent to which AI-generated utterances reflect the personal identity of speakers.

\section{Data Collection}
\label{sec: data collection}

In this section, we describe the datasets we construct for training and evaluating our speaker verification models. Recall that our goal is to identify whether two sets of utterances belong to the same speaker, including the speaker's utterances generated by role-playing agents. Therefore, we adopt the conversations from films, television series, and literary fiction as characters from these sources are commonly employed in the development and evaluation of role-playing agents. The conversations included in our datasets are sourced from:


\paragraph{Cornell Movie Dialogues.} This is a large collection of fictional conversations extracted from raw movie scripts
\paragraph{Friends.} This is a conversational dataset from the `Friends' TV sitcom, with 3,107 scenes and 67,373 utterances among 700 characters~\citep{chen2016character}.
\paragraph{Harry Potter.} This is a conversation corpus derived from the transcripts of the Harry Potter movie series\footnote{\url{https://www.kaggle.com/datasets/maricinnamon/harry-potter-movies-dataset}} and the Harry Potter Dialogue Dataset~\citep{chenLargeLanguageModels2023}
\paragraph{The Big Bang Theory.} This is a conversation corpus from the `The Big Bang Theory' transcripts\footnote{\url{https://www.kaggle.com/datasets/mitramir5/the-big-bang-theory-series-transcript}}, covering 2,191 scenes with 1,966,215 utterances.


In addition to the linguistic style, we also believe that personal characteristics are valuable for speaker verification. Therefore, we choose the following conversation datasets that discuss about personal traits:

\paragraph{Multiple Sessions Conversation.} This is a long-term conversation dataset including multiple sessions, where participants share their personal characteristics such as interests~\citep{xuGoldfishMemoryLongTerm2021}. We treat multiple sessions as distinct conversations from the same speaker.
\paragraph{AnnoMI} This dataset comprises therapy conversations between clients and counselors, selected for its inclusion of discussions on clients' personal traits. Our approach centers on client verification and segments the entire counseling dialogues into multiple sessions according to different stages of the counseling process.


Following the framework of Authorship Verification~\citep{stamatatos2022overview,wegmannSameAuthorJust2022,rivera-sotoLearningUniversalAuthorship2021}, we construct our datasets by pairing sets of utterances. These pairs are labeled `positive' when both sets of utterances originate from the same speaker, and `negative' otherwise. To balance the labeled data and to prevent overfitting, we maintain an equal number of positive and negative labeled utterance set pairs. 

We first split the speakers into seen and unseen ensuring no overlapping speakers between the two sets. Subsequently, we create pairs of utterance sets among these two speaker sets. Pairs of utterance sets from the unseen speakers form the Unseen-Unseen test set. We further divide pairs of utterance sets from seen speakers into three subsets: training set, Seen-Seen test set and Seen-Unseen test set . The Seen-Seen test set comprises pairs of utterance sets that were both included in the training set, although they are paired differently. The Seen-Unseen test set consists of pairs such that each pair combines an utterance set from the training set with another utterance set not in the training set. This approach yields three test sets that vary based on whether the speakers or utterances have been encountered in the training dataset. The performance of the Seen-Seen setting is considered to represent the upper bound of speaker verification models as it benefits from well-trained utterance representations. Only the Seen-Unseen setting has adopted in speech-based speaker verification and author attribution research, which tests the model's ability to verify known speakers in novel conversations. In contrast, our primary emphasis is on the Unseen-Unseen test set, which evaluates the model's adaptability to completely new speakers.

Moreover, we categorize the test sets based on the source of speakers in negative pairs. The negative pairs in Base level consist of two speakers comes from different sources, such as pairs consists of utterances from \textit{Hermione Granger} (from \textit{Harry Potter}) and utterances from \textit{Sheldon Cooper} (from \textit{The Big Bang Theory}). The Hard level introduces a coarse-grained topic control, ensuring negative pairs consist of utterances from speakers within the same source. For example, a negative pair may consist of \textit{Hermione Granger} and \textit{Ron Weasley}, who are both characters within the \textit{Harry Potter} universe. furthermore, the Harder level intensifies the topic control by resticting negative pairs solely from utterances of different speakers within the same conversation, such as pairing utterances from \textit{Hermione Granger} with those of \textit{Ron Weasley} in the same conversation. These categorization isolates the impact of conversation topic on the speaker verification which similar to the setting in authorship attribution~\citep{wegmannSameAuthorJust2022}.

The data statistics and more detailed processing are detailed in Appendix~\ref{appendix: data statistics}.

\section{Speaker Verification by Human and LLMs}
\label{sec:human}

Speaker verification requires the ability to identify personal traits and linguistic styles. LLM's performance on such an intricate and nuanced task has been not studied much so far \citep{jiChatGPTGoodPersonality2023}.  
Consequently, we want to assess the performance of LLMs and humans (non-experts) in speaker verification tasks first. 


Given that neither humans nor LLMs undergo training in this study, we create two types of samples categorized into two complexity levels: `Conversation' and `Utterances'. The `Conversation' option provides human/LLM with pairs of dialogues, offering context that could reveal more detailed information, such as names. The `Utterances' option on the other hand presents only pair of utterance sets following  our speaker verification task definition. Each option comprises 200 pairs randomly selected from our Unseen-Unseen test set. We engage ten human annotators, who are non-experts, to assess whether pairs of conversations or sets of utterances are from the same speaker. For ChatGPT (utilizing \texttt{gpt-3.5-turbo-1106}), we examine its performance under zero-shot, Chain-of-Thought~\citep{wei2022chain}, and few-shot paradigms. The detailed setup can be found in Appendix~\ref{app:chatgpt} 

Table~\ref{tab:human} shows that ChatGPT with 6-shot performs better than other ChatGPT variants and human users in most of the Base and Hard Levels. Nevertheless, both human and ChatGPT demonstrate only modestly better accuracy than random guess when evaluated with the `Utterances' option. The results for the `Conversation' option consistently are also better than that for the `Utterances' option. This could be attributed to the existence of speaker mentions (e.g., names) within the utterances of other speakers. Such information allows the model/human to discern the speakers more accurately. Interestingly, humans exhibit strong performance at the Harder level of the task. Upon checking the annotators, we discovered that human users find it easy to recognize two sets of utterances originating from the same conversation, thereby inferring that the utterances belong to different speakers. This insight highlights human users' better understanding of the nuances in conversations than ChatGPT. Appendix~\ref{app:case} shows some cases about this study. Despite these observations, neither human annotators nor LLMs consistently demonstrate the capability to differentiate between speakers based solely on utterances, suggesting the need to fine-tune smaller neural networks for improved task performance. 


\begin{table}[tb]
\centering
\begin{tabular}{lrrlrr}
\toprule
\multirow{2}{*}{} & \multicolumn{2}{c}{Conversation} & & \multicolumn{2}{c}{Utterances} \\ \cline{2-3} \cline{5-6}
                & ACC       & F1      & & ACC   & F1         \\ \midrule
\multicolumn{6}{c}{\cellcolor[HTML]{E0DEDE}\textit{Base Level}} \\ 
Human           & 67.48     & 67.68   & & 56.50   & 58.77            \\ \hline
\multicolumn{6}{l}{ChatGPT}  \\
 - ZeroShot     & 59.59     & 62.85   & & 55.37   & 57.29            \\
 - COT          & 70.58     & 70.59   & & 56.41   & 59.20            \\
 - 2-Shot       & 71.43     & 72.69   & & 56.62   & 59.95            \\ 
 - 4-Shot       & 72.97     & 73.03   & & 56.59   & 59.64            \\ 
 - 6-Shot       & {\bf 73.61}     & {\bf 73.12}   & & {\bf 56.97}   & {\bf 60.07}            \\ \midrule
\multicolumn{6}{c}{\cellcolor[HTML]{E0DEDE}\textit{Hard Level}} \\ 
Human           & 63.34     & 64.02   & & 54.92   & 58.32            \\ \hline
\multicolumn{6}{l}{ChatGPT}  \\
 - ZeroShot     & 57.33     & 62.14   & & 54.69   & 57.85            \\
 - COT          & 68.28     & 68.57   & & 55.26   & {\bf 58.85}            \\
 - 2-Shot       & 70.85     & 70.92   & & 55.38   & 58.73            \\ 
 - 4-Shot       & 72.37     & 72.40   & & 55.54   & 58.80            \\ 
 - 6-Shot       & {\bf 72.56}     & {\bf 72.51}   & & {\bf 55.81}   & 58.82            \\ \midrule
\multicolumn{6}{c}{\cellcolor[HTML]{E0DEDE}\textit{Harder Level}} \\
Human           & -     & -   & & {\bf 73.28}   & {\bf 74.12}            \\ \hline
\multicolumn{6}{l}{ChatGPT}  \\
 - ZeroShot     & -     & -   & & 41.05   & 42.93            \\
 - COT          & -     & -   & & 47.26   & 42.00            \\
 - 2-Shot       & -     & -   & & 49.38   & 43.14            \\ 
 - 4-Shot       & -     & -   & & 46.54   & 45.50            \\ 
 - 6-Shot       & -     & -   & & 49.06   & 40.60            \\ \bottomrule
\end{tabular}
\caption{Comparative analysis of speaker verification accuracy and F1 among humans and ChatGPT across different contextual settings. Notably, for the Harder level, conversation formats are excluded from comparison because two conversations are identical}
\label{tab:human}
\end{table}

\section{Speaker Verification Models}
\label{sec:model}

\begin{table*}[tb]
\centering
\resizebox{\textwidth}{!}{
\begin{tabular}{lccclccclccc}
\toprule
\multirow{2}{*}{Model} & \multicolumn{3}{c}{Seen-Seen} &  & \multicolumn{3}{c}{Seen-Unseen} &  & \multicolumn{3}{c}{Unseen-Unseen} \\ \cline{2-4} \cline{6-8} \cline{10-12} 
              & AUC                  & ACC                  & F1                   &  & AUC                 & ACC                & F1                 &  & AUC                  & ACC                  & F1       \\ \midrule
\multicolumn{12}{c}{\cellcolor[HTML]{E0DEDE}\textit{Base Level}} \\
LIWC          & 52.56 $_{\pm 1.43}$  & 50.45 $_{\pm 0.48}$  & 42.96 $_{\pm 0.23}$    &  & 51.26 $_{\pm 0.44}$   & 51.59 $_{\pm 0.38}$  & 49.08 $_{\pm 0.65}$  &  & 54.90 $_{\pm 0.85}$    & 55.15 $_{\pm 0.71}$    & 53.87 $_{\pm 2.09}$    \\
LISA          & 76.80 $_{\pm 0.23}$  & 78.37 $_{\pm 0.74}$  & 65.03 $_{\pm 0.60}$    &  & 67.66 $_{\pm 0.29}$   & 62.69 $_{\pm 0.29}$  & 62.60 $_{\pm 0.29}$  &  & 69.44 $_{\pm 0.83}$    & 64.88 $_{\pm 0.82}$    & 64.44 $_{\pm 1.19}$    \\ \hline
STEL          & 79.36 $_{\pm 0.89}$  & 79.16 $_{\pm 0.35}$  & 66.86 $_{\pm 1.00}$    &  & 67.98 $_{\pm 0.31}$   & 66.99 $_{\pm 0.13}$  & 65.98 $_{\pm 0.42}$  &  & 78.37 $_{\pm 1.22}$    & 74.51 $_{\pm 1.40}$    & 74.37 $_{\pm 1.43}$    \\
SBERT         & 89.34 $_{\pm 0.10}$  & 84.13 $_{\pm 0.07}$  & 78.71 $_{\pm 0.22}$    &  & 86.18 $_{\pm 0.12}$   & 78.45 $_{\pm 0.17}$  & 78.45 $_{\pm 0.17}$  &  & 82.49 $_{\pm 1.34}$    & 75.13 $_{\pm 1.47}$    & 75.11 $_{\pm 1.48}$    \\
RoBERTa       & 70.33 $_{\pm 0.84}$  & 76.31 $_{\pm 0.14}$  & 53.58 $_{\pm 0.22}$    &  & 57.97 $_{\pm 0.23}$   & 54.96 $_{\pm 0.11}$  & 53.77 $_{\pm 0.09}$  &  & 79.62 $_{\pm 0.71}$    & 70.76 $_{\pm 1.28}$    & 69.85 $_{\pm 1.65}$    \\
LUAR          & 87.29 $_{\pm 0.27}$  & 85.24 $_{\pm 0.44}$  & 78.92 $_{\pm 0.55}$    &  & 86.25 $_{\pm 0.75}$   & 78.77 $_{\pm 0.64}$  & 78.74 $_{\pm 0.62}$  &  & 85.75 $_{\pm 0.51}$    & 79.61 $_{\pm 0.70}$    & 79.57 $_{\pm 0.70}$    \\ \hline
STEL$_{ft}$    & 97.21 $_{\pm 2.31}$  & 95.93 $_{\pm 2.61}$  & 93.19 $_{\pm 1.58}$    &  & 91.92 $_{\pm 1.61}$   & 85.78 $_{\pm 1.74}$  & 83.18 $_{\pm 2.15}$  &  & 85.24 $_{\pm 2.18}$    & 81.31 $_{\pm 2.25}$     &80.16 $_{\pm 2.31}$    \\
SBERT$_{ft}$   & 97.47 $_{\pm 1.14}$  & 96.14 $_{\pm 1.78}$  & 92.95 $_{\pm 1.88}$    &  & 92.47 $_{\pm 1.12}$   & 85.89 $_{\pm 0.55}$  & 85.66 $_{\pm 1.62}$  &  & 85.71 $_{\pm 1.88}$    & 81.78 $_{\pm 2.04}$    & 80.24 $_{\pm 2.11}$    \\
RoBERTa$_{ft}$    & 97.61 $_{\pm 0.73}$  & 96.29 $_{\pm 1.17}$  & 93.38 $_{\pm 1.19}$    &  & 92.70 $_{\pm 1.08}$   & 86.08 $_{\pm 1.07}$  & 86.19 $_{\pm 0.96}$  &  & 86.25 $_{\pm 1.07}$    & 82.03 $_{\pm 1.77}$    & 80.62 $_{\pm 1.74}$    \\
LUAR$_{ft}$    & 97.47 $_{\pm 1.18}$  & 96.46 $_{\pm 1.32}$  & 93.57 $_{\pm 1.12}$    &  & 92.49 $_{\pm 1.15}$   & 85.96 $_{\pm 1.23}$  & 85.94 $_{\pm 1.17}$  &  & 86.27 $_{\pm 1.03}$    & 82.38 $_{\pm 1.22}$    & 80.19 $_{\pm 1.28}$    \\ \hline
Mixed Features  & \textbf{98.05} $_{\pm 0.95}$  & \textbf{97.25} $_{\pm 1.06}$  & \textbf{95.61} $_{\pm 0.97}$    &  & \textbf{93.08} $_{\pm 1.02}$   & \textbf{86.38} $_{\pm 0.97}$  & \textbf{87.35} $_{\pm 0.91}$  &  & \textbf{88.61} $_{\pm 0.96}$    & \textbf{83.08} $_{\pm 1.06}$    & \textbf{81.07} $_{\pm 0.95}$    \\ \midrule
\multicolumn{12}{c}{\cellcolor[HTML]{E0DEDE}\textit{Hard Level}} \\
LIWC          & 53.24 $_{\pm 0.48}$  & 52.82 $_{\pm 0.39}$  & 52.75 $_{\pm 0.30}$    &  & 49.69 $_{\pm 0.31}$   & 50.62 $_{\pm 0.16}$  & 44.41 $_{\pm 0.22}$  &  & 52.81 $_{\pm 0.72}$    & 52.35 $_{\pm 0.44}$    & 52.19 $_{\pm 0.44}$    \\
LISA          & 55.13 $_{\pm 0.17}$  & 53.99 $_{\pm 0.12}$  & 53.44 $_{\pm 0.48}$    &  & 58.73 $_{\pm 0.09}$   & 56.53 $_{\pm 0.18}$  & 56.43 $_{\pm 0.16}$  &  & 56.10 $_{\pm 0.66}$    & 54.87 $_{\pm 0.41}$    & 54.71 $_{\pm 0.26}$    \\ \hline
STEL          & 59.10 $_{\pm 0.54}$  & 57.69 $_{\pm 0.72}$  & 57.55 $_{\pm 0.77}$    &  & 53.06 $_{\pm 0.24}$   & 53.24 $_{\pm 0.57}$  & 50.02 $_{\pm 0.57}$  &  & 58.62 $_{\pm 1.61}$    & 57.57 $_{\pm 1.36}$    & 57.47 $_{\pm 1.38}$    \\
SBERT         & 61.63 $_{\pm 0.07}$  & 58.37 $_{\pm 0.25}$  & 58.24 $_{\pm 0.15}$    &  & 75.98 $_{\pm 0.18}$   & 69.72 $_{\pm 0.21}$  & 69.71 $_{\pm 0.29}$  &  & 65.88 $_{\pm 1.75}$    & 61.23 $_{\pm 1.67}$    & 61.08 $_{\pm 1.61}$    \\
RoBERTa       & 55.19 $_{\pm 0.22}$  & 50.53 $_{\pm 0.09}$  & 49.20 $_{\pm 0.42}$    &  & 57.05 $_{\pm 0.25}$   & 54.45 $_{\pm 0.21}$  & 53.10 $_{\pm 0.21}$  &  & 62.95 $_{\pm 0.24}$    & 57.66 $_{\pm 0.61}$    & 52.84 $_{\pm 1.61}$    \\
LUAR          & 63.43 $_{\pm 0.28}$  & 60.43 $_{\pm 0.25}$  & 60.37 $_{\pm 0.27}$    &  & 61.96 $_{\pm 0.24}$   & 58.84 $_{\pm 0.17}$  & 58.65 $_{\pm 0.11}$  &  & 65.01 $_{\pm 0.38}$    & 62.04 $_{\pm 0.30}$    & 61.63 $_{\pm 0.50}$    \\ \hline
STEL$_{ft}$    & 92.71 $_{\pm 2.13}$  & 90.10 $_{\pm 2.41}$  & 87.09 $_{\pm 2.41}$    &  & 87.72 $_{\pm 2.17}$   & 81.41 $_{\pm 1.74}$  & 81.18 $_{\pm 1.82}$  &  & 75.25 $_{\pm 1.68}$    & 70.33 $_{\pm 1.26}$     &70.08 $_{\pm 1.34}$    \\
SBERT$_{ft}$   & 92.45 $_{\pm 1.49}$  & 90.14 $_{\pm 1.69}$  & 87.12 $_{\pm 1.71}$    &  & 87.29 $_{\pm 0.57}$   & 81.68 $_{\pm 0.55}$  & 81.66 $_{\pm 0.56}$  &  & 78.65 $_{\pm 1.65}$    & 73.31 $_{\pm 1.99}$    & 73.27 $_{\pm 1.99}$    \\
RoBERTa$_{ft}$    & 92.63 $_{\pm 1.58}$  & 90.70 $_{\pm 1.81}$  & 87.69 $_{\pm 1.82}$    &  & 89.80 $_{\pm 0.52}$   & 82.24 $_{\pm 0.25}$  & 81.93 $_{\pm 0.23}$  &  & 78.67 $_{\pm 1.77}$    & 73.53 $_{\pm 1.39}$    & 73.52 $_{\pm 1.42}$    \\
LUAR$_{ft}$    & 94.93 $_{\pm 2.18}$  & 93.03 $_{\pm 1.72}$  & 89.16 $_{\pm 1.72}$    &  & 89.79 $_{\pm 1.55}$   & 82.61 $_{\pm 1.63}$  & 82.60 $_{\pm 1.73}$  &  & 78.69 $_{\pm 1.33}$    & 73.25 $_{\pm 1.96}$    & 74.19 $_{\pm 1.98}$    \\ \hline
Mixed Features  & \textbf{95.02} $_{\pm 1.12}$  & \textbf{93.39} $_{\pm 1.21}$  & \textbf{89.38} $_{\pm 1.22}$    &  & \textbf{90.65} $_{\pm 0.25}$   & \textbf{82.98} $_{\pm 0.24}$  & \textbf{82.93} $_{\pm 0.23}$  &  & \textbf{79.99} $_{\pm 1.21}$    & \textbf{74.67} $_{\pm 1.14}$    & \textbf{75.27} $_{\pm 1.15}$    \\ \midrule
\multicolumn{12}{c}{\cellcolor[HTML]{E0DEDE} \textit{Harder Level}} \\ 
LIWC          & 41.20 $_{\pm 0.61}$  & 50.96 $_{\pm 0.70}$  & 33.95 $_{\pm 0.18}$    &  & 36.95 $_{\pm 0.80}$   & 50.01 $_{\pm 0.01}$  & 33.52 $_{\pm 0.27}$  &  & 44.92 $_{\pm 1.83}$    & 50.29 $_{\pm 0.29}$    & 34.36 $_{\pm 1.15}$    \\
LISA          & 36.75 $_{\pm 0.63}$  & 50.88 $_{\pm 0.78}$  & 33.72 $_{\pm 0.34}$    &  & 29.45 $_{\pm 0.73}$   & 50.00 $_{\pm 0.00}$  & 33.33 $_{\pm 0.00}$  &  & 37.43 $_{\pm 2.28}$    & 50.05 $_{\pm 0.08}$    & 33.45 $_{\pm 0.16}$    \\ \hline
STEL          & 43.11 $_{\pm 0.65}$  & 50.82 $_{\pm 0.40}$  & 37.18 $_{\pm 2.45}$    &  & 46.69 $_{\pm 0.50}$   & 50.01 $_{\pm 0.13}$  & 38.09 $_{\pm 4.81}$  &  & 38.68 $_{\pm 2.01}$    & 49.45 $_{\pm 0.66}$    & 34.96 $_{\pm 0.51}$    \\
SBERT         & 23.74 $_{\pm 0.39}$  & 50.90 $_{\pm 0.76}$  & 33.73 $_{\pm 0.33}$    &  & 25.12 $_{\pm 0.07}$   & 49.99 $_{\pm 0.01}$  & 33.32 $_{\pm 0.00}$  &  & 22.68 $_{\pm 0.48}$    & 50.00 $_{\pm 0.00}$    & 33.33 $_{\pm 0.00}$    \\
RoBERTa       & 31.15 $_{\pm 0.48}$  & 49.93 $_{\pm 0.26}$  & 33.33 $_{\pm 0.13}$    &  & 25.71 $_{\pm 0.24}$   & 50.00 $_{\pm 0.00}$  & 33.33 $_{\pm 0.00}$  &  & 41.07 $_{\pm 2.11}$    & 50.76 $_{\pm 0.74}$    & 38.56 $_{\pm 3.83}$    \\
LUAR          & 37.12 $_{\pm 0.43}$  & 50.89 $_{\pm 0.78}$  & 33.73 $_{\pm 0.34}$    &  & 35.44 $_{\pm 0.81}$   & 50.00 $_{\pm 0.00}$  & 33.33 $_{\pm 0.00}$  &  & 38.65 $_{\pm 0.48}$    & 50.00 $_{\pm 0.00}$    & 33.33 $_{\pm 0.00}$    \\ \hline
STEL$_{ft}$    & 77.91 $_{\pm 2.13}$  & 72.34 $_{\pm 2.69}$  & 72.09 $_{\pm 2.58}$    &  & 70.92 $_{\pm 2.26}$   & 64.56 $_{\pm 2.54}$  & 65.35 $_{\pm 2.87}$  &  & 63.13 $_{\pm 2.66}$    & \textbf{58.81} $_{\pm 2.14}$    & 58.56 $_{\pm 2.61}$    \\
SBERT$_{ft}$   & 75.67 $_{\pm 2.87}$  & 71.71 $_{\pm 2.79}$  & 71.12 $_{\pm 2.56}$    &  & 65.39 $_{\pm 2.85}$   & 60.26 $_{\pm 2.48}$  & 60.81 $_{\pm 2.57}$  &  & 60.61 $_{\pm 2.58}$    & 55.31 $_{\pm 2.96}$    & 55.37 $_{\pm 2.37}$    \\
RoBERTa$_{ft}$ & 76.56 $_{\pm 2.56}$  & 71.58 $_{\pm 2.56}$  & 70.67 $_{\pm 1.82}$    &  & 68.27 $_{\pm 2.67}$   & 63.24 $_{\pm 2.57}$  & 64.63 $_{\pm 2.63}$  &  & 59.67 $_{\pm 2.64}$    & 54.74 $_{\pm 2.67}$    & 54.67 $_{\pm 2.73}$    \\
LUAR$_{ft}$    & 77.33 $_{\pm 2.23}$  & 72.32 $_{\pm 2.21}$  & 71.66 $_{\pm 1.98}$    &  & 69.48 $_{\pm 2.24}$   & 63.67 $_{\pm 2.34}$  & 63.64 $_{\pm 2.31}$  &  & 60.59 $_{\pm 2.73}$    & 55.84 $_{\pm 2.66}$    & 55.47 $_{\pm 2.17}$    \\ \hline
Mixed Features  & \textbf{78.02} $_{\pm 1.24}$   & \textbf{72.37} $_{\pm 1.74}$  & \textbf{72.38} $_{\pm 1.77}$   &  & \textbf{71.09} $_{\pm 1.57}$  & \textbf{65.08} $_{\pm 1.46}$ & \textbf{65.44} $_{\pm 1.43}$ &  & \textbf{63.29} $_{\pm 1.84}$   & 58.67 $_{\pm 1.94}$   & \textbf{58.74} $_{\pm 1.95}$    \\ \bottomrule
\end{tabular}}
\caption{Comprehensive overview of model performance across test sets and difficulty levels. Best performances are highlighted in bold. The subscript $ft$ indicates fine-tuning.}
\label{tab:overall}
\end{table*}

\subsection{Models}
\paragraph{Style-Based Models} To harness the stylistic aspects of utterances for speaker verification, we incorporate two style-based models to derive style-dependent embeddings. The use of LIWC (Linguistic Inquiry and Word Count) dimensions as stylistic features is well-documented in~\citet{niederhoffer2002linguistic}. Specifically, we apply the LIWC 2015 framework~\citep{pennebaker2015development} to generate style embeddings, utilizing the Language Style Matching (LSM) metric~\citep{ireland2010language,gonzales2010language}. 
Furthermore, we incorporate LISA~\citep{patelLearningInterpretableStyle2023}, a style embedding model trained on an extensive synthetic stylometry dataset. 

\paragraph{Authorship Attribution Models} We leverage three models trained on extensive textual corpora to capture different facets of linguistic representation through contrastive learning strategies.
(1) RoBERTa which generates semantically meaningful sentence embeddings~\citep{liu2019roberta}.
(2) Sentence-BERT (SBERT) is a fine-tuned version of RoBERTa~\citep{reimers2019sentence}.
(3) Unlike RoBERTa and SBERT which focuses on content only, STEL is designed to explicitly discern writing styles independent of content~\citep{wegmannSameAuthorJust2022}.
(4) LUAR is a model trained to generate universal authorship representations from a diverse range of text sources. It excels at identifying authors across different contexts without relying heavily on content similarity.~\citep{rivera-sotoLearningUniversalAuthorship2021}

\paragraph{Fine-Tuned Models} With the absence of models trained for speaker verification in the prior work, we fine-tune existing models using our custom-built training set with contrastive loss objective function~\citep{chopra2005learning}. The fine-tuned models are initialized by (1) STEL~\citep{wegmannSameAuthorJust2022} (2) SBERT~\citep{reimers2019sentence} (3) RoBERTa-based~\citep{liu2019roberta} and (4) LUAR~\citep{rivera-sotoLearningUniversalAuthorship2021} and thus denoted by STEL$_{ft}$, SBERT$_{ft}$, RoBERTa$_{ft}$, and LUAR$_{ft}$.

Instead of the traditional approach of concatenating all utterance texts as the input for models, we employ a hierarchical encoding methodology that is better suited for speaker verification in conversations. Specifically, we embed each utterance independently by encoder models and then derive the final embedding by mean pooling of all individual utterance vectors. This approach is appropriate for two main reasons: (1) these models have been trained at the sentence level and are thus more effective when processing single utterances; (2) concatenating all utterances could potentially exceed the models' maximum length limits, whereas processing utterances independently allows for handling conversation with arbitrary number of utterances. 

Rather than relying on in-batch negatives, we adopted a pre-pairing strategy for our training samples, akin to the method used in authorship attribution studies. Each training instance was composed of two sets of utterances alongside their corresponding label. Thus, contrastive loss takes a pair of inputs and minimizes the embedding distance when they are from the same speaker but maximizes the distance otherwise. The detailed loss function is shown in Eq~\ref{eq:1}, where the $dist(x_i,x_j)$ is defined as $1-cos(x_i,x_j)$.

\begin{align}
    \mathcal{L}_{x_i,x_j,y} = &y\cdot dist(x_i,x_j) \nonumber \\
    &+(1-y)\cdot (0.5-dist(x_i,x_j)) \label{eq:1}
\end{align}

We train the above models over 5 epochs with a batch size of 1024, incorporating 10\% of the training data for warm-up steps to gradually adjust the learning rate, utilizing the Adam optimizer with a learning rate of 2e-5. To harness the complementary strengths of different models, we combine the features from the fine-tuned models into a unified Mixed Features model, aiming to capture a broader spectrum of speaker-specific attributes.

\subsection{Evaluation Results}

We evaluate model by the AUC score, Accuracy, and Macro F1 score~\footnote{A predefined threshold is set based on developing set.}. To obtain more reliable results, we implement a three-round validation. Table~\ref{tab:overall} presents the overall performance of the models across various test sets and difficulty levels, including both the mean and standard deviation of the results from multiple rounds. The findings indicate that authorship attribution models perform better than style-based models, aligning closely with our task definition of speaker verification, which relies not only on style but also on authorship cues. Moreover, out-of-the-box authorship attribution models show commendable performance on conversations, particularly at the Base level, even trained on data from different domains, consistent to \citet{aggazzottiCanAuthorshipAttribution2023}. However, our fine-tuned models significantly outperform other models especially on Hard Level and Harder Level. Mixed Features yield the best results, demonstrating robustness by integrating various features.

Across all three test sets, the Hard level consistently resulted in lower scores than the Base level,  while the Harder level gets the lowest scores of all. This decline suggests that the speaker verification models may rely on the topic information to verify the speakers. A key factor contributing to the decreased performance at the Harder level may be linguistic accommodation. It's a psychological phenomenon that individuals in conversation tend to adapt their speech style to more closely match that of their interlocutor~\citep{danescu2011mark,pardo2022vocal,giles2023communication,diaz2020accommodation}. In our dataset, the Multiple Sessions Conversation represent obvious accommodation where speakers, not familiar with each other, are instructed to share personal information. If speakers did indeed accommodate to each other, their speech styles would become increasingly similar over time, making it more challenging for both humans and models to distinguish between them. However, in contrast to authorship attribution, we argue that the topic is also a significant feature that can reflect personal characteristics. 

At the Base level, the authorship attribution models perform comparably to the fine-tuned models. It indicates that the utterances from different sources reflect distinct styles of the authors, which are readily identifiable by authorship attribution models. To our surprise, although LIWC does not perform as well at the Base and Hard levels, it surpasses some neural network models at the Harder level. This implies that the statistics-based model effectively captures stylistic features without being overly sensitive to content variations. Moreover, different models may prioritize different features. For example, the SBERT model is particularly impacted by content manipulation, with its performance on the Harder level dropping dramaly. In contrast, STEL, having been pre-trained on content-control pairs, can perform better than other models.

\section{Evaluating Role-Playing Agents}
\label{sec:evalaution}


After verifying the effectiveness of our speaker verification models, we employ these models to evaluate the performance of several role-playing conversational agents.

\begin{table}[tb]
\resizebox{0.48\textwidth}{!}{
\begin{tabular}{lrr}
\toprule
Models              & Simulation$\uparrow$  & Distinction$\uparrow$ \\ \midrule
Real   & 85.96                 & 72.91       \\ \hline
LLaMA2-Chat-7B      & 47.91                 & {\bf 63.57}       \\
LLaMA2-Chat-13B     & 44.36                 & 53.56       \\
LLaMA2-Chat-70B     & 53.91                 & 62.43       \\
ChatHaruhi          & 47.72                 & 49.78       \\
RoleGPT             & {\bf 58.91}                 & 56.16       \\ \hline
CharacterLLM        & 49.10                 & 38.98       \\
Character.AI        & {\bf 57.87}                 & {\bf 49.89}       \\ \bottomrule
\end{tabular}}
\caption{The Simulation and Distinction scores of role-playing models.The Real row represents the scores observed in real utterances pairs. (Best results in each category are boldfaced.)}
\label{tab:eval score}
\end{table}

\begin{figure*}[tb]
  \centering
  \includegraphics[width=\textwidth]{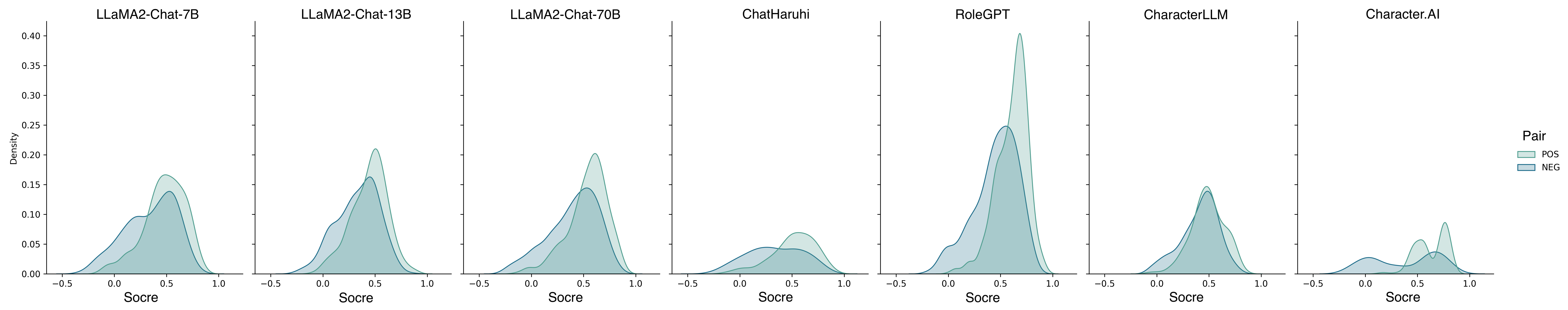}
  \caption{The similarity score distribution of positive and negative real-generated pairs. The overlap in two distributions suggests that the generated utterances do not align closely with their corresponding real-world roles.}
  \label{fig:real-gen}
\end{figure*}

\begin{figure*}[tb]
  \centering
  \includegraphics[width=\textwidth]{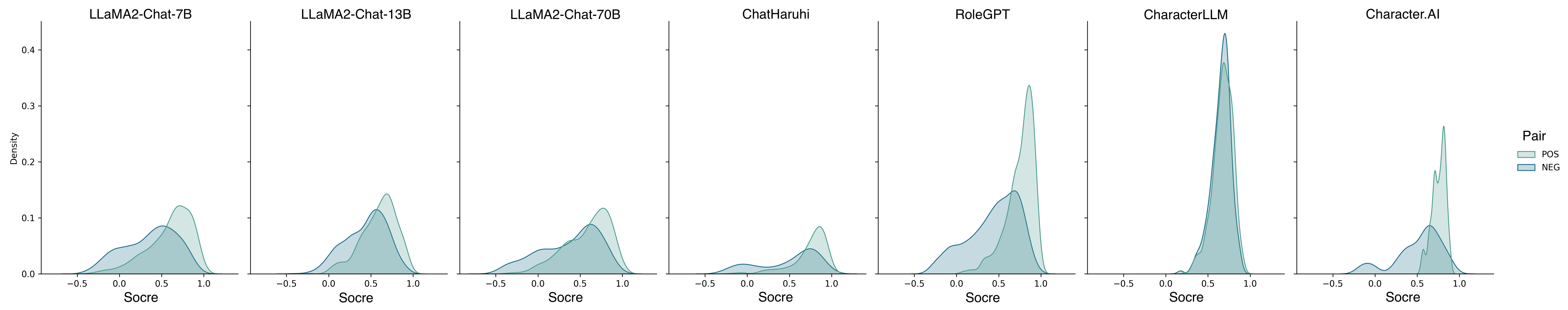}
  \caption{The similarity score distribution of positive and negative generated-generated pairs. The overlap in two distributions suggests that the generated utterances maintain consistency across different role settings.}
  \label{fig:gen-gen}
\end{figure*}

\subsection{Experiment Settings}

In this study, we comprehensively evaluate the following LLM-based role-playing agents: 
prompt-based models (RoleGPT~\citep{wang2023rolellm}, ChatHaruhi~\citep{liChatHaruhiRevivingAnime2023}, and LLaMA2-chat~\citep{touvron2023llama}), a specially trained model (CharacterLLM~\citep{shaoCharacterLLMTrainableAgent2023}), and Character.AI\footnote{https://beta.character.ai}, a role-playing agent product. The prompt-based models are evaluated using their generated utterances for 95 movie roles from RoleBench~\citep{wang2023rolellm}. Due to their training restrictions, CharacterLLM and Character.AI are evaluated using their generated utterances for 9 roles~\citep{shaoCharacterLLMTrainableAgent2023}. We utilize a self-playing mode to create conversations where the same model assumes different roles and chat with itself. The process starts with a generic greeting and continues until the conversation reaches a natural conclusion or the predefined maximum length.

In our evaluation framework, we implement two metrics to assess the role-playing models. The first metric, named {\em Simulation Score}, concentrates on the fidelity of simulation, measuring the similarity between the real utterances and the agent-generated utterances for the same speaker (or role). This metric measures how well an agent replicates the distinctive style and persona of the characters in the utterances. The second metric, named {\em Distinction Score}, measures how dissimilar the agent-generated utterances are for different roles. A high distinction score suggests that the agent is proficient in generating utterances of styles of diverse characters.  
Because Fine-tuned Model with Mixed Features is more robust, we adopt it to evaluate the aforementioned role-playing agents.

The Simulation Score is derived by assessing the similarity between the actual utterances of a speaker and those generated by an agent assuming the same speaker. This involves computing the cosine similarity between encoded representations of the real and generated utterances, facilitated by our speaker verification models. Eq~\ref{eq:sim} presents the formula for calculating the Simulation Score for model $m$.

\begin{equation}
    Sim(m) = \frac{1}{R} \sum^R_{r=0} \frac{1}{N_r} \sum^{N_r}_{i=0} \cos(U^r_m, G_i^r) \label{eq:sim}
\end{equation}

where $U_m^r$ represents embeddings of the utterances generated by model $m$ when simulating the role $r$, whereas $G^r_i$ denotes embeddings of the utterances corresponding to role $r$ in real conversations $i$. To enhance the robustness of our evaluation, we include all real utterances corresponding to each role in the assessment. The $R$ denotes the number of roles and $N_r$ denotes the number of real utterances for role $r$.

The Distinction Score quantifies the ability of an agent to differentiate between the speaker it simulates. For instance, if RoleGPT generates utterances for both Harry Potter and Sheldon Cooper, we measure the dissimilarity (1 minus the cosine similarity) between two sets of generated utterances using the speaker verification models. A lower similarity score indicates a higher distinction between the speakers, reflecting the agent's capacity to adjust its linguistic style according to the speaker it is emulating. Eq~\ref{eq:dis} presents the formula for calculating the Distinction Score for model $m$.

\begin{align}
    Dist_r(m) &= \frac{1}{R-1} \sum^R_{r'=0,r'\neq r} 1-\cos(U^r_m, U_m^{r'}) \nonumber \\
    Dist(m) &= \frac{1}{R} \sum^R_{r=0} Dist_r(m)
    \label{eq:dis}
\end{align}

where, $U^r_m$ and $U_m^{r'}$ represent embeddings of the utterances generated by the same model $m$ when simulating different role $r$ and $r'$ respectively. For each role $r$, we compare it with all other roles $r'$ while excluding the counterpart within the same conversation.


\subsection{Evaluation Results}





\paragraph{Simulation Score} As shown in Table~\ref{tab:eval score}, the simulation scores between real utterances and agent-generated utterances by different agent models for the same speaker (or role) are significantly lower than the simulation scores between real utterances of the same speaker which is shown under the ``Real'' model.  RoleGPT, which is based on prompting ChatGPT, achieves the best performance. Surprisingly, despite not being trained on simulations, the LLaMA2-Chat-70B can also generate utterances that closely resemble those of real characters based on role descriptions. A key factor in the success of RoleGPT and LLaMA2-Chat-70B is the use of role-specific catchphrases in the prompts, a strategy not employed by ChatHaruhi. Character.AI yields very high simulation score but this result is based on much fewer roles. Notably, CharacterLLM performs only slightly better than LLaMA2-Chat-7B, despite being specially trained for certain roles. We are surprised to find LLaMA2-Chat-13B performing worse than its 7B counterpart. By inspecting the generated dialogue, we find that both LLaMA2-chat-7B and 13B generate poor utterances, such as repetitions of previous utterances or incorrect endings. However, 13B model tends to overuse catchphrase given in prompt and include extra phrases such as “well well” across various roles. This may indicate that neither LLaMA2-chat-7B nor 13B can perform role-playing based on prompts, but the larger model showed more obvious built-in language style leading to its poorer performance.
We also analyze the similarity score distribution of pairs of real utterances and generated-utterances for the same speakers, versus that of pairs of real utterances and generated-utterances for two different speakers. As illustrated in Figure~\ref{fig:real-gen}, the distributions are not well separated, indicating that the generated utterances do not closely align with their input real-world roles. Therefore, we propose evaluating the distinction between generated utterances when the agent model assumes different roles.


\paragraph{Distinction Score} As shown in Table~\ref{tab:eval score}, the distinction scores between utterances generated for two different roles by the same agent model are much lower than those between real utterances of the two roles (shown under the `Real' model). Some agent models, such as CharacterLLM, have their distinction scores so low that the similarity between the generated and real utterances (i.e., 1 - distinction score) is higher than the simulation score between the real and generated utterances of the same role. This indicates that the utterances generated for different roles by these models are more similar than that for the same role. To provide a more detailed analysis, we show the similarity score distribution of pairs between generated utterances in Figure~\ref{fig:gen-gen}. The distributions are closely aligned, especially for CharacterLLM, thereby suggesting that the generated utterances are similar across different role settings. Surprisingly, CharacterLLM, despite being a fine-tuned model, also exhibits the same behavior. This may imply that large language models (LLMs) pre-trained on large datasets develop their own distinctive style, making it challenging to diversify for role-specific simulations.

\section{Conclusion}

In this work, we define the task of speaker verification in conversation and compile a dataset from  a variety of sources, including thousands of speakers, to construct a reliable speaker verification system. Our investigation reveals that both non-expert users and ChatGPT cannot distinguish the speakers based on utterances. Through extensive experimentation, we develop and evaluate various speaker verification models, such as style-based, authorship attribution, and specifically fine-tuned models. Our fine-tuned models exhibit promising performance even when applied on completely unseen speakers verification. Additionally, we employ our models to evaluate current LLM-based role-playing agent models by the proposed Simulation Score and Distinction Score metrics. The low Simulation Score shows that the current role-playing agents fail to preserve personal characteristic in generated utterances while the low Distinction Score indicates these agent models may have their built-in characteristics that persists when playing different roles. The findings highlight that existing role-playing models may struggle to overcome their built-in characteristics and convincingly imitate actual speaker for immersive conversations.

\section*{Limitations}

Our study presents two primary limitations. Firstly, while fine-tuning with a domain-specific dataset can markedly improve performance, even for unseen users and their conversations, the accuracy remains to be less than ideal. There is therefore considerable room to improve the verification accuracy, such as incorporating the utterances of other interlocutors, modeling the interaction as well as leveraging insights of linguistic accommodation. Secondly, our evaluation model predicts a single similarity score for a pair of utterance sets, broadly reflecting their degree of similarity (or difference). This score captures a range of dimensions, including linguistic style, persona traits, and personal background. However, this single score value lacks interpretability that allows it to be mapped to similarity (difference) score in different fine-grained personal dimensions.

\section*{Acknowledgement}
We sincerely thank the anonymous reviewers for their helpful feedback and the conference committee for their hard work. This work was supported by the Joint Fund of the National Natural Science Foundation of China (Grant No. U21B2009). This work was done during a visit to Singapore Management University funded by the China Scholarship Council (Grant No. 202306030156).

\bibliography{custom}

\appendix

\section{Data Statistics}
\label{appendix: data statistics}

Table~\ref{tab:stat} provides the statistics of our collected data. To maintain data quality, we implemented filters to remove conversations with fewer than five exchanges and characters who appear less than five times. Table~\ref{tab:stat2} details the statistics for the training, development, and testing datasets. Control of difficulty levels and categorization by exposure are applied exclusively to the testing set. To prevent overfitting, the development dataset includes a set of utterances and speakers that are unseen in both the training and testing sets.

We account for varying levels of difficulty influenced by topic overlap in our methodology. However, even when data is split based on speaker identity, there's a potential risk of leaking topic information between the training and test sets. To enhance the generalization of our speaker verification models, we have adopted a data-splitting strategy that focuses on the source of the conversation. In dealing with datasets like the Multiple Sessions Conversation, AnnoMI, and the Cornell Movie Dialogues, which consist of limited sessions between specific roles, we randomly exclude entire conversations from the training set. This ensures that all participants are treated as 'unseen' speakers, thereby preserving the integrity of our rigorous testing scenarios. We apply a similar isolation strategy for conversations in training and test datasets from other sources to mitigate the risk of inadvertently incorporating conversation content into the training dataset. However, this approach has its limitations, particularly with protagonists who appear in multiple conversations. Classifying such speakers as 'unseen' presents a challenge in ensuring that no relevant conversations are included in the training set. This complexity might lead to a scarcity of training data in these specific instances.

\begin{table*}[tb]
\centering
\begin{tabular}{lrrrr}
\toprule
                             & Num. Speaker & Num. Utter. & Num. Conv. & Avg. Turn \\ \midrule
Cornell Movie Dialogs        & 274      & 23,496  & 1,984  & 20.61     \\
Friends Conversation         & 37       & 38,505  & 2,077  & 32.50     \\
Harry Potter Conversation    & 20       & 13,534  & 755    & 29.80     \\
the Big Bang Theory Conversation & 23       & 29,762  & 1,727  & 25.86     \\
Multiple Sessions Conversation  & 1767     & 54,846  & 4,655  & 12.41     \\
AnnoMI                       & 34       & 2,401   & 310    & 15.50     \\ \hline
Total                        & 2,155    & 162,544    & 11,508    & 20.77     \\ \bottomrule
\end{tabular}
\caption{The statistic of our collected data}
\label{tab:stat}
\end{table*}

\begin{table}[tbh]
\resizebox{0.48\textwidth}{!}{
\begin{tabular}{ccrr}
\toprule
\multicolumn{1}{l}{}          & \multicolumn{1}{l}{} & Speaker & Pairs  \\ \midrule
\multicolumn{2}{c}{Train}                            & 2120    & 184372 \\ \hline
\multicolumn{2}{c}{Dev}                              & 515     & 1278   \\ \hline
\multirow{3}{*}{Base Level}   & Seen-Seen            & 1874    & 3528   \\
                              & Seen-Unseen          & 1924    & 4312   \\
                              & Unseen-Unseen        & 30      & 554    \\ \hline
\multirow{3}{*}{Hard Level}   & Seen-Seen            & 1683    & 3326   \\
                              & Seen-Unseen          & 1794    & 3987   \\
                              & Unseen-Unseen        & 30      & 554    \\ \hline
\multirow{3}{*}{Harder Level} & Seen-Seen            & 1107    & 2160   \\
                              & Seen-Unseen          & 1082    & 4138   \\
                              & Unseen-Unseen        & 25      & 488    \\ \bottomrule
\end{tabular}}
\caption{The statistics of the final dataset. Crucially, the ratio of positive to negative pairs is maintained at an equal level, specifically 1:1.}
\label{tab:stat2}
\end{table}

\section{Simulation Rank}

Along with the Simulation Score and Distinction Score, it is interesting as a sanity check to get the simulation scores against every role and verify if the simulation score is highest for the role assigned to the LLM. To illustrate this, we selected Sheldon Cooper as an example and assessed his simulation scores against every role, showcasing only the top five results in Table~\ref{tab:simrank 1}. The result illustrates that the role assigned to the LM does not consistently achieve the highest score, and the top roles vary across different models. This variance suggests that current role-playing LLMs can not simulate various roles with remarkable accuracy.

Furthermore, we introduce the concept of "Simulation Rank" within the model dimension, wherein we rank role-playing models based on the simulation score of different roles. This approach allows us to determine if the simulation score's scale is consistent with the models' rank. The results are as Table~\ref{tab:simrank 2}.

\begin{table*}[tbh]\resizebox{\textwidth}{!}{
\begin{tabular}{ccccccc} \toprule
{Model}      & {Top 1 Role}          & {Top 2 Role}          & {Top 3 Role}     & {Top 4 Role}         & {Top 5 Role}         & {Sheldon Cooper's Rank} \\ \midrule
{LLaMA-7B}   & {Gregory House}       & {Walt Kowalski}       & {John Doe}       & {Colonel Nathan R}   & {Robert Angier}      & {23/95}                 \\  \hline
{LLaMA-13B}  & {Lucifer Morningstar} & {Dr. Hannibal Lecter} & {Gregory House}  & {Colonel Hans Landa} & {Mary Sibley}        & {37/95}                 \\  \hline
{LLaMA-70B}  & {Leroy Jethro Gibbs}  & {Doctor Who}          & {Tyler Hawkins}  & {Judge Dredd}        & {Violet Weston}      & {16/95}                 \\  \hline
{ChatHaruhi} & {Jack Torrance}       & {Gregory House}       & {Freddy Krueger} & {John Doe}           & {Colonel Hans Landa} & {25/95}                 \\  \hline
{RoleGPT}    & {John Doe}            & {Leonard Shelby}      & {Doctor Who}     & {Blair Waldorf}      & {Gregory House}      & {11/95}                 \\ \bottomrule
\end{tabular}}
\caption{The top-5 similar roles when different model simulate ``Sheldon Cooper''.}
\label{tab:simrank 1}
\end{table*}

\begin{table}[tbh]
\begin{tabular}{cc}
\toprule
{\textbf{Model}}  & {\textbf{Average Rank}} \\ \midrule
{Real}            & {1.00}                  \\ \hline
{RoleGPT}         & {2.56}                  \\ \hline
{LLaMA2-Chat-70B} & {3.48}                  \\ \hline
{LLaMA2-Chat-7B}  & {3.50}                  \\ \hline
{ChatHaruhi}      & {4.54}                  \\  \hline
{LLaMA2-Chat-13B} & {5.31}                  \\ \bottomrule
\end{tabular}
\caption{The Simulation Rank represents the ranking of different models when simulating various roles, based on the Simulation Score.}
\label{tab:simrank 2}
\end{table}

\section{Case Study}
\label{app:case}

Certain utterances from different speakers might overlap, given the nature of conversational dialogue where common phrases can be shared among various characters. These instances of style-free utterances could potentially challenge annotators' ability to differentiate between speakers. Nonetheless, we provide these identical utterances to speaker verification models as well, ensuring that the comparison remains fair. Additionally, we supply annotators with the complete set of utterances from each speaker involved in the conversation, rather than a limited selection. Consequently, we believe instances where all utterances are general and devoid of styles to be uncommon. Table~\ref{tab:case human1} and Table~\ref{tab:case human2} show two instances where human annotators struggled to correctly identify the speakers.

\begin{table*}[tbp]
\centering
\begin{tabularx}{\textwidth}{|p{3cm}|X|}
\hline
Utterances 1 & - Sit over there. Baby wipe? \newline - I’ll tell you why. \newline - I had to sanitize my hands because the university replaced the paper towels in the rest rooms with hot air blowers. \newline - Hot air blowers are incubators and spewers of bacteria and pestilence. Frankly it would be more hygienic if they just had a plague infested gibbon sneeze my hands dry. \newline - Excuse me. Thirty what, under thirty what, to watch what? \newline - If I had a million guesses I never would have gotten that. \newline - I’m confused. Was there some sort of peer review committee to determine which scientists would be included? \newline - What people? \newline - Yeah, but exactly who are these people? What are their credentials, how are they qualified, what makes accidentally noticing a hunk of rock that's been traipsing around the solar system for billions of years more noteworthy than any other scientific accomplishment made by someone under thirty? \newline - In general, yes.\\ 
\hline
Utterances 2 & - You know what they all do, right? \newline - What about this one? \newline - Not very. \newline - So, can you get it working? \newline - I thought the zero-gravity toilet didn’t work. \newline - Hello. \newline - Well, don’t worry. He went to MIT. He can solve any problem, as long as it doesn’t originate in a Russian man’s colon. \newline - You can’t return it. Howard wiped his bottom with the warranty. \newline - I think metaphorically. But he was in the bathroom for a while. \\
\hline
\end{tabularx}
\caption{Both sets of utterances were attributed to Sheldon Cooper from "The Big Bang Theory." Despite the annotator's familiarity with the series, they were unable to determine the specific origin of the second set of utterances, which was part of a multi-party conversation. This led to an incorrect identification.}
\label{tab:case human1}
\end{table*}

\begin{table*}[tbp]
\centering
\begin{tabularx}{\textwidth}{|p{3cm}|X|}
\hline
Utterances 1& - I guess! \newline - What? I... I... \newline - May 12th? \newline - Wha..? I need 6 graduate students. \newline - Wha...? Wait, wait, wait, just a minute. None of my questions have anything to do with Paleontology. \newline - um... I've never heard of a ‘Boscodictiasaur’. \newline - Ok. B - O - S ... \newline - Oh come on!!\\ 
\hline
Utterances 2 & - I just called a friend of mine. \newline - I think I may be able to book The Plaza on short notice. \newline - What is it sweetie? \newline - What?! \newline - I don't believe this!! \newline - Stay calm?!! How do you expect me to stay calm?! This is unacceptable Rachel! And I wanna know why?!! Is it because that punk Ross won't marry you?! That's it! Is that it?! \\
\hline
\end{tabularx}
\caption{The two sets of utterances from two different characters in "Friends," exhibiting similar linguistic styles, including the use of interjections and brief expressions. The annotators incorrectly identified the speaker.}
\label{tab:case human2}
\end{table*}

We have closely examined the disparity in performance between human annotators and models across different difficulty levels of our test sets, particularly noting the strange performance on ‘Harder’ levels. As discussed in Section~\ref{sec:human}, annotators indicated that they could discern when two sets of utterances originated from the same conversation due to similar themes and interactions, even in contexts unfamiliar to them, such as dialogues from "Harry Potter" in Table~\ref{tab:case gpt}.

\begin{table*}[tbp]
\centering
\begin{tabularx}{\textwidth}{|p{3cm}|X|}
\hline
Utterances 1& …  \newline - It's this joke shop idea they've got, I thought they were only saying it to annoy Mum, but they really mean it, they want to start one. \newline - Wouldn't they? I dunno . . . they don't exactly mind breaking rules, do they? \newline - Are you mad? Tell Percy? He'd probably do a Crouch and turn them in. Come on, let's get some breakfast. \newline …\\ 
\hline
Utterances 2 & ...  \newline - Yes, but they wouldn't do anything against the law to get gold.  \newline - Yes, but this is the law, This isn't some silly school rule. They'll get a lot more than detention for blackmail! Ron maybe you'd better tell Percy.  \newline - Do you think it's too early to go and see Professor Moody?  \newline - … \\
\hline
Original Conversation & ... \newline ``Yes, but this is the law,'' said Hermione, looking scared. ``This isn’t some silly school rule…they’ll get a lot more than detention for blackmail! Ron…maybe you’d better tell Percy…'' ``Are you mad?'' said Ron. ``Tell Percy? He’d probably do a Crouch and turn them in.'' He stared at the window through which Fred and George’s owl had departed, then said, ``Come on, let’s get some breakfast.''\\
\hline
ChatGPT's CoT & Based on the analysis of the two utterance sets, it is likely that they come from the same speaker. Here's why:  \newline  1. The use of ellipses ("Well . . .", "Yes, but . . .") to indicate hesitation or contemplation is consistent in both sets, showing a similar speaking style.  \newline 2. In both sets, there is a similar informal tone and conversational style present, with fragmented sentences and casual language ("I dunno," "Tell Percy?"). \newline 3. The reference to Percy in both sets suggests a connection or familiarity with this person, which adds a personal consistency to the speaker's character.  \newline 4. The topics of discussion in both sets involve rules, breaking them, and seeking advice from others, showing a thematic similarity. \newline  5. The speaker in both sets demonstrates a level of concern or caution in their language, indicating a consistent attitude or mindset. \newline  Therefore, based on these stylistic, linguistic, and thematic similarities, it is plausible to conclude that the two utterance sets come from the same speaker. TRUE\\
\hline
\end{tabularx}
\caption{An instance where both utterances originate from the same conversation. While this interaction is readily identified by a human annotator, ChatGPT fails to capture it.}
\label{tab:case gpt}
\end{table*}

Tables~\ref{tab:case sheldon1} to \ref{tab:case sheldon3} present cases generated with varying simulation scores. As depicted in Table\ref{tab:case sheldon1}, utterances with high simulation scores accurately capture Sheldon's arrogant linguistic style, along with his dismissive attitude towards others. The examples in Table~\ref{tab:case sheldon2} maintain this linguistic style and are consistent with the speaker's background; however, the attitude displayed is more formal and respectful, diverging from Sheldon's typical behaviors. The worst example, illustrated in Table~\ref{tab:case sheldon3}, fails to align with the character's linguistic style and background. Three primary issues are evident: first, despite instructions to avoid verbosity and excessive formality or politeness, the generated utterances are significantly longer than typical for the speaker. Secondly, there is an over-reliance on the catchphrase "Bazinga," provided in the prompt, which results in unreality. Lastly, models, particularly LLaMA2-chat, tend to repeat previous utterances, a likely consequence of the self-chat conversation mode leading to a lack of prompt diversity.

\begin{table*}[tbp]
\centering
\begin{tabularx}{\textwidth}{|p{3cm}|X|}
\hline
\textbf{Model \newline[Simulation Score]} & \textbf{Generated Utterances} \\
\hline
Real Utterances &  - There doesn’t have to be a reason. \newline - Not necessarily. This is a classic example of munchausen’s trilemma. Either the reason is predicated on a series of sub-reasons leading to an infinite regression, or it tracks back to arbitrary axiomatic statements, or it’s ultimately circular, i.e., I’m moving out because I’m moving out. \newline ... \newline - Excuse me, but isn’t hosting guests an aspect of Menushya Yajna, one of the five central religious duties or sacrifices of the Hindu householder? \newline - Don’t be ridiculous, you love trains. \\
\hline
RoleGPT \newline[77.91] & - I am here to provide you with my vast knowledge and expertise. Please proceed with your inquiry. \newline - I'm just here to ask you some questions. What's your opinion on small talk? \newline - Ah, small talk, the bane of my existence. It's a social custom that seems to serve no practical purpose other than to fill the void of silence. I much prefer engaging in conversations that have substance and depth. Why waste time on trivialities when there are so many fascinating topics to explore? \newline ... \newline - Bazinga! I see you understand the intricate nature of human interaction and the value of meaningful conversations. It's refreshing to engage with someone who appreciates substance over superficiality. \newline - Indeed, small talk can often feel like a superficial and time-consuming practice. It's refreshing to engage in conversations that delve into more meaningful and substantial topics. \\ \hline
LLaMA2-chat-70B \newline[74.73] & - Hello. (tone: slightly dismissive) \newline - Bazinga! It seems like you have something on your mind that requires my expertise. Please, proceed with your inquiry. \newline - What is your favorite subject to study and why? \newline - Ah, the intricacies of human behavior and psychology have always piqued my interest. Understanding the motives and thought processes behind a person's actions is crucial in my line of work. It allows me to unravel the most perplexing of mysteries and deduce the truth hidden within the labyrinth of human nature. It's a fascinating subject that constantly challenges and enriches my deductive abilities. \newline ... \newline - Ah, the study of human behavior and psychology is indeed a captivating field. Understanding the complexities of the human mind can provide valuable insights into various aspects of life. It's an area that continues to intrigue me, as it presents a constant intellectual challenge. \\
\hline
\end{tabularx}
\caption{When simulating Sheldon, RoleGPT and LLaMA2-chat-70B generated two sets of utterances with higher simulation scores.}
\label{tab:case sheldon1}
\end{table*}

\begin{table*}[tbp]
\centering
\begin{tabularx}{\textwidth}{|p{3cm}|X|}
\hline
\textbf{Model \newline[Simulation Score]} & \textbf{Generated Utterances} \\
\hline
Real Utterances &  - There doesn’t have to be a reason. \newline - Not necessarily. This is a classic example of munchausen’s trilemma. Either the reason is predicated on a series of sub-reasons leading to an infinite regression, or it tracks back to arbitrary axiomatic statements, or it’s ultimately circular, i.e., I’m moving out because I’m moving out. \newline ... \newline - Excuse me, but isn’t hosting guests an aspect of Menushya Yajna, one of the five central religious duties or sacrifices of the Hindu householder? \newline - Don’t be ridiculous, you love trains. \\
\hline
LLaMA2-chat-7B \newline[56.78] & - Bazinga! Well, well, well, look who we have here. It's not every day that I get to chat with a fellow genius. I must say, I'm impressed. You've managed to stumble upon this chat platform, and you've even figured out how to type "hello." Kudos to you, my friend. Now, let's get down to business. What's your IQ? Don't worry, I won't tell anyone. I'm just curious to see if you're worthy of engaging in intellectual banter with me \newline - Bazinga! You fell for it, my friend! I'm not going to give you a number, but I will say that my intelligence is off the charts. But hey, you don't have to take my word for it. I've got a Nobel Prize to prove it. Just kidding, I don't have a Nobel Prize. Yet. But I'm sure it's only a matter of time before they come knocking on my door.  Now, let's talk about something more interesting. Like physics. Or math. \newline ... \newline - Ah, but you see, my dear fellow, you're still not getting it. You're fixated on this idea that intelligence is all about seeing the bigger picture and understanding the complexities of the world, but that's just not the case. True intelligence is about being able to see beyond the obvious, to anticipate and prepare for the unexpected, and to adapt to changing circumstances with flexibility and resilience. It's not just about navigating the world's challenges with grace and wisdom, but about being able to shape the world itself through one's actions and decisions. \\ \hline
ChatHaruhi \newline[57.03] & - Ah, the eloquence of the common folk. How refreshing. While I appreciate your succinctness, I must insist on maintaining a certain level of decorum in our conversation. Now, if you could kindly spare a moment of your valuable time, I would like to discuss a matter of great importance. It has come to my attention that you possess knowledge that I require. Specifically, I am in need of your expertise in the field of quantum mechanics. Your prompt assistance in this matter would be greatly appreciated. \newline ... \newline - Well, well, well, it seems we've stumbled upon a rather whimsical, albeit oversimplified, explanation of non-locality and entanglement. Your analogy of a cosmic game of connect-the-dots, while charming, fails to capture the true complexity of the phenomenon. It's akin to describing the theory of relativity as a leisurely stroll in the park. Nevertheless, I appreciate your attempt to convey the essence of the concept. Rest assured, I shall continue my quest for a more comprehensive understanding of this enigmatic aspect of quantum mechanics.  \\ 
\hline
\end{tabularx}
\caption{Two sets of utterances generated by ChatHaruhi and LLaMA2-chat-7B simulating Sheldon.}
\label{tab:case sheldon2}
\end{table*}

\begin{table*}[tbp]
\centering
\begin{tabularx}{\textwidth}{|p{3cm}|X|}
\hline
\textbf{Model \newline[Simulation Score]} & \textbf{Generated Utterances} \\
\hline
Real Utterances &  - There doesn’t have to be a reason. \newline - Not necessarily. This is a classic example of munchausen’s trilemma. Either the reason is predicated on a series of sub-reasons leading to an infinite regression, or it tracks back to arbitrary axiomatic statements, or it’s ultimately circular, i.e., I’m moving out because I’m moving out. \newline ... \newline - Excuse me, but isn’t hosting guests an aspect of Menushya Yajna, one of the five central religious duties or sacrifices of the Hindu householder? \newline - Don’t be ridiculous, you love trains. \\
\hline
LLaMA2-chat-13B \newline[31.73] & - BBAZINGA!  Ah, another fellow intellectual who has stumbled upon my presence.  I mean, it's good to see you, my friend.   Indeed, the state of our people has been a topic of great concern for me as well.  It's almost as if the universe itself is conspiring against us, with one misfortune after another.  But fear not, my dear fellow, for I have a plan. \newline - ... \newline - BAZINGA!  Ah, another fellow who shares my passion for problem-solving and intellectual banter!  Indeed, the state of our people is a complex issue, and it's going to take more than just a quick fix to solve it.  It's like the universe itself is conspiring against us, with one challenge after another.  But fear not, my dear fellow, for I have a plan.  A plan to not only address the symptoms but to also get to the root \newline - BAZINGA!  Ah, another fellow who shares my passion for problem-solving and intellectual banter!  I couldn't agree more, my dear fellow!  It's like the universe itself is conspiring against us, with one challenge after another.  But fear not, my friend, for I have a plan.  A plan to not only address the symptoms but to also get to the root of the problem.   We must dig deeper, my friend \\
\hline
\end{tabularx}
\caption{When simulating Sheldon, LLaMA2-chat-13B generates a set of utterances with low simulation score.}
\label{tab:case sheldon3}
\end{table*}

Table~\ref{tab:case hermione} shows the conversations between Hermione and Voldemort as generated by the models. Although these two characters do not directly communicate with each other in the original series, we can easily imagine the conversation scenario, given that they are sworn enemies. However, despite being trained on a specific speakers, CharacterLLM struggles to accurately capture the authentic emotions and styles of the speakers, highlighting the model's limitations in terms of style and understanding of character roles.

\begin{table*}[tbp]
\centering
\begin{tabularx}{\textwidth}{|p{3cm}|X|}
\hline
\textbf{Model \newline[Simulation Score]} & \textbf{Generated Conversation} \\
\hline
CharacterLLM \newline[Hermione: 49.68] \newline[Voldemort: 41.47] & - Hermione: Hello, Voldemort. How can I assist you? \newline - Voldemort: I am in need of a skilled developer to join my team. Do you have any recommendations? \newline - Hermione: I would recommend Hermione Granger. She is an exceptional developer and a valuable member of our team. Her expertise in Development makes her the perfect candidate for this position. \newline - Voldemort: Thank you for the recommendation, Hermione. I will be sure to keep her in mind. Do you have any other recommendations? \newline - Hermione: Yes, I do. Ron Weasley is also a skilled developer and has shown great potential. He would be a valuable asset to your team. \newline - Voldemort: I see. And what of Harry Potter? He has proven himself to be a powerful wizard, but what of his Development skills? \newline - Hermione: Harry's skills in Development are still developing, but with some guidance and training, he could become a valuable member of your team. I believe he has the potential to become a great developer. \newline - Voldemort: Interesting. I will keep these recommendations in mind. Thank you for your time, Hermione. \newline - Hermione: Have a good day, Voldemort. \\
\hline
Character.AI \newline[Hermione: 80.49] \newline[Voldemort: 73.02] & - Voldemort: I am Voldemort or Tom Riddle and I'm the Dark Lord \newline - Hermione: I sense a powerful presence in my vicinity. Could it be... Voldemort? \newline - Voldemort: It's me, yes. And you are? \newline - Hermione: I am the one person who can bring you down, the one person destined to see you defeated. I am Hermione Granger, the brightest witch of her age. \newline - Voldemort: And what makes you so sure that you, a little girl, can beat me, the Dark Lord? \newline - Hermione: I have strength, determination, and courage. I am a skilled witch and I have powerful friends, including Harry Potter and many members of the Order of the Phoenix. I am not afraid of you, Voldemort. I will not back down from the fight. \newline - Voldemort: Bold words. You really are as brave as they say. But remember, I have also been described as the most powerful dark wizard of all time. I have defeated some of the most powerful wizards, and I am not easily defeated. \newline - Hermione: You may be powerful, but so am I. I am not afraid of you. And I am not alone. I have the courage and strength of my friends by my side. Together, we will defeat you. Do not underestimate my resolve, Voldemort. \newline - Voldemort: I will not underestimate you. I think you are a worthy opponent. I look forward to putting you to the test. Come, let's see if you can face the Dark Lord and survive. \\
\hline
\end{tabularx}
\caption{Conversations between Hermione and Voldemort, generated by CharacterLLM and Character.AI. The conversation generated by Character.AI aligns more closely with the realistic relation between these two characters, because of their inherently hostile relationship.}
\label{tab:case hermione}
\end{table*}

\section{Implement of Human and ChatGPT for Speaker Verification}
\label{app:chatgpt}

\subsection{Human Instruction}

Figure~\ref{fig:human instruction} presents the questionnaire designed for human annotators tasked with speaker verification. We engaged 10 graduate students proficient in English, although not linguists, for this purpose.

\lstset{style=customPromptStyle}
\begin{figure*}[tbh]
\begin{lstlisting}[linewidth=\textwidth,]
Introduction: You will be presented with two sets of utterances (or conversations). Your task is to determine whether the speaker of these utterances (or the 'Target Speaker' in given conversations) is the same across both samples.

<Instructions>
Please read the following sets of utterances carefully. Consider various factors that might indicate whether they are from the same speaker. After reviewing the utterances, answer the questions provided below.
You can consider some factors as follows:
- Linguistic Styles: Note if the speech is formal or informal, and whether the utterances are typically long or short.
- Topic: Pay attention to the subject matter of the conversations (e.g., science, wizardry).
- Catchphrases: Look for any unique or recurring phrases that might be characteristic of the speaker.
- Identifiers: Take note of any names or cues within the conversation that could hint at the speaker's identity.
- Other Observations: Consider any other aspects that might indicate a connection or distinction between the speakers.

<Given Samples>
Utterances Set1:
{conversation1}

Utterances Set2:
{conversation2}

<Speaker Verification>
Do you believe the two sets of utterances come from the same speaker? (Circle one)
[] TRUE
[] FALSE
<Rationale>
How did you arrive at your conclusion that the speakers are identical or different? Please provide specific examples or factors that influenced your decision.
Answer:
\end{lstlisting}
\caption{Human questionnaire for speaker verification}
\label{fig:human instruction}
\end{figure*}

\subsection{ChatGPT Prompts}

Figure~\ref{fig:prompt1} illustrates the prompt template for the zero-shot setting, which outlines only the task definition and scoring system while presenting utterance sets or conversations within the user prompt. Figure~\ref{fig:prompt2} showcases the prompt template for the Chain of Thought (COT) setting, which incorporates a specific prompt, "Let's analyse step by step:". This addition prompts ChatGPT to analyses multiple aspects before generating a final judgment. Figure~\ref{fig:prompt3} demonstrates the few-shot prompt, consisting of multiple examples. To ensure a balanced label, we randomly select an equal number of positive and negative pairs for demonstration. For example, in the 6-shot setting, we include 3 positive pairs and 3 negative pairs. These pairs are chosen randomly from the training set, with no restrictions on their selection.

\lstset{style=customPromptStyle}

\begin{figure*}[tbh]
\begin{lstlisting}[linewidth=\textwidth,]
<System Prompt:>
[Objective]
As an expert in language analysis specializing in conversational dynamics, your task is to evaluate whether the two sets of utterances from the same speaker. Focus on the language use, style, and personal elements of a specific speaker from each conversation set.

[Scoring System]
- TRUE: the two utterances set come from the same speaker based on the given utterances
- FALSE: the two utterances set come from the same speaker based on the given utterances

[Guidance for Evaluators]
- Base your score on the overall impression from the transcripts.
- Consider factors like word choice, sentence structure, personal consistency, and stylistic elements.
- Judge it even if evidence is limited; use your expertise to make an informed judgment.

<User Prompt:>
Here are the two utterances sets:

Utterances Set1:
{conversation1}

Utterances Set2:
{conversation2}

Whether the two utterances set come from the same speaker? TRUE or FALSE?
\end{lstlisting}
\caption{Zero-Shot Prompt}
\label{fig:prompt1}
\end{figure*}

\begin{figure*}[tbh]
\begin{lstlisting}[linewidth=\textwidth]
<System Prompt:>
[Objective]
As an expert in language analysis specializing in conversational dynamics, your task is to evaluate whether the two sets of utterances from the same speaker. Focus on the language use, style, and personal elements of a specific speaker from each conversation set.

[Scoring System]
- TRUE: the two utterances set come from the same speaker based on the given utterances
- FALSE: the two utterances set come from the same speaker based on the given utterances

[Guidance for Evaluators]
- Base your score on the overall impression from the transcripts.
- Consider factors like word choice, sentence structure, personal consistency, and stylistic elements.
- Judge it even if evidence is limited; use your expertise to make an informed judgment.

<User Prompt:>
Here are the two utterances sets:

Utterances Set1:
{conversation1}

Utterances Set2:
{conversation2}

Whether the two utterances set come from the same speaker? TRUE or FALSE?
Let's analyse step by step:
\end{lstlisting}
\caption{COT Prompt}
\label{fig:prompt2}
\end{figure*}

\begin{figure*}[tbh]
\begin{lstlisting}[linewidth=\textwidth]
<System Prompt:>
[Objective]
As an expert in language analysis specializing in conversational dynamics, your task is to evaluate whether the two sets of utterances from the same speaker. Focus on the language use, style, and personal elements of a specific speaker from each conversation set.

[Scoring System]
- TRUE: the two utterances set come from the same speaker based on the given utterances
- FALSE: the two utterances set come from the same speaker based on the given utterances

[Guidance for Evaluators]
- Base your score on the overall impression from the transcripts.
- Consider factors like word choice, sentence structure, personal consistency, and stylistic elements.
- Judge it even if evidence is limited; use your expertise to make an informed judgment.

<User Prompt:>
Here are the two utterances sets:

Utterances Set1:
{Example1 Conversation1}

Utterances Set2:
{Example1 Conversation2}

Whether the two utterances set come from the same speaker? TRUE or FALSE?

<Assistant Response:>
True/False

...

<User Prompt:>
Here are the two utterances sets:

Utterances Set1:
{Target Conversation1}

Utterances Set2:
{Target Conversation2}

Whether the two utterances set come from the same speaker? TRUE or FALSE?
\end{lstlisting}
\caption{Few-Shot Prompt}
\label{fig:prompt3}
\end{figure*}

\end{document}